\documentclass[letterpaper, 10 pt, conference]{ieeeconf}

\IEEEoverridecommandlockouts                              %
\usepackage[T1]{fontenc}
\usepackage[utf8]{inputenc}

\usepackage{amsmath} %
\usepackage{amssymb} %

\usepackage{hyperref}
\usepackage{graphicx}

\usepackage{multirow}
\usepackage{booktabs}

\usepackage{cleveref}

\usepackage{bm}
\usepackage{url}

\DeclareMathOperator*{\argmax}{arg\,max}

\usepackage{ifthen}

\title{\LARGE \bf
Matched Filtering based LiDAR Place Recognition\\for Urban and Natural Environments
}

\author{Therese Joseph\hskip5em Tobias Fischer\hskip5em Michael Milford%
\thanks{The authors are with the QUT Centre for Robotics, School of Electrical Engineering and Robotics, Queensland University of Technology, Brisbane, QLD 4000, Australia. Email: {\tt\footnotesize t2.joseph@hdr.qut.edu.au}}%
\thanks{This work received funding from ARC Laureate Fellowship FL210100156 to MM, and from a grant from Intel Labs to TF and MM. The authors acknowledge continued support from the Queensland University of Technology (QUT) through the Centre for Robotics.}%
}

\begin{document}
\maketitle
\thispagestyle{empty}
\pagestyle{empty}

\begin{abstract}

Place recognition is an important task within autonomous navigation, involving the re-identification of previously visited locations from an initial traverse. Unlike visual place recognition (VPR), LiDAR place recognition (LPR) is tolerant to changes in lighting, seasons, and textures, leading to high performance on benchmark datasets from structured urban environments. However, there is a growing need for methods that can operate in diverse environments with high performance and minimal training. In this paper, we propose a handcrafted matching strategy that performs roto-translation invariant place recognition and relative pose estimation for both urban and unstructured natural environments. Our approach constructs Birds Eye View (BEV) global descriptors and employs a two-stage search using matched filtering --- a signal processing technique for detecting known signals amidst noise. Extensive testing on the NCLT, Oxford Radar, and WildPlaces datasets consistently demonstrates state-of-the-art (SoTA) performance across place recognition and relative pose estimation metrics, with up to 15\% higher recall than previous SoTA.

\end{abstract}

\section{Introduction}

\begin{figure}[t]
    \centering
    \includegraphics[width=0.99\columnwidth]{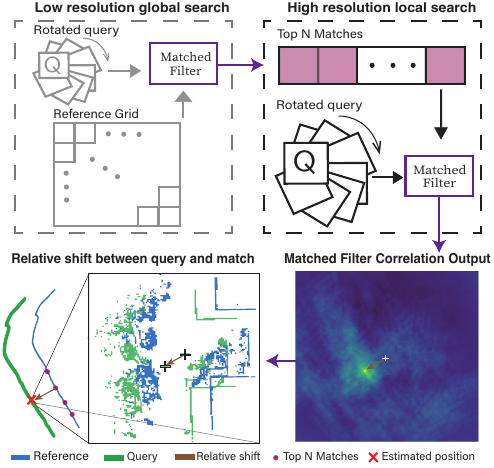}
    \vspace{-0.4cm}
    \caption{Matched filter-based LiDAR Place Recognition (LPR) architecture. Low-resolution reference BEV descriptors are accumulated within a grid while the query BEV descriptor is rotated for a global search using the matched filter. The top $n$ matches from this search are then used for a high-resolution local search. The resulting correlation output from the top match is used for translation estimation and pose correction from reference to query traverse. }
    \label{fig:1.1}
    \vspace{-0.6cm}
\end{figure}

In autonomous navigation, place recognition is the task of global relocalisation given a reference set of previously visited places. This capability is crucial for several downstream tasks in Simultaneous Localisation and Mapping (SLAM), such as loop closure detection and localisation in GPS-denied environments \cite{lowry2015visual}, \cite{garg2021your}. Much of the research has focused on visual place recognition (VPR), which relies on features extracted from vision-based sensors to distinguish different locations, as demonstrated in methods like FAB-MAP \cite{cummins2008fab}, SeqSLAM \cite{milford2012seqslam}, and NetVLAD \cite{arandjelovic2016netvlad}. Since visual features are sensitive to changes in lighting, weather and viewpoints; VPR methods rely on diverse training data that captures varied conditions to handle these changes. In contrast, geometric features captured by LiDAR sensors are inherently tolerant to variations in lighting, weather and seasons. Consequently, the field has seen the emergence of LiDAR Place Recognition (LPR), which utilises handcrafted or learned methods to transform long-range point cloud data with 360$^\circ$ field of view into efficient representations of a place \cite{komorowski2022improving}, \cite{kim2018scan}, \cite{chen2021overlapnet}.

While trained LPR methods can leverage large datasets to generate robust, efficient solutions, they often suffer from performance degradation in novel, unseen environments \cite{knights2023geoadapt}. On the other hand, handcrafted methods, which do not require extensive training data, may struggle in diverse environments depending on the design of descriptors and tuning of parameters. The ideal LPR system should consistently perform well across varied environments with minimal training and fine-tuning. Motivated by this challenge, we propose a robust handcrafted matching strategy from signal processing called matched filtering -- where detecting a known signal in the presence of noise closely parallels the place recognition problem. By leveraging a handcrafted Bird's Eye View (BEV) descriptor, combined with a global and local search architecture, our method effectively matches previously visited places with the current query, even in noisy and variable conditions as illustrated in Figure \ref{fig:1.1}.

Another challenge for LPR methods is achieving both rotational and translational invariance when positional discrepancies exist between the reference and query traverse. While some methods address this by generating roto-translation (rotation and translation) invariant descriptors \cite{xu2023ring++}, \cite{he2016m2dp}, \cite{liu2019lpd}, this invariance often comes at the cost of losing the capability to directly estimate relative pose -- a critical downstream task in applications like loop closure in SLAM. To bridge this gap, some LPR methods apply point cloud registration techniques like ICP \cite{besl1992method} and RANSAC \cite{fischler1981random} after place recognition to refine the scan alignment and estimate relative pose \cite{xu2023ring++}, \cite{luo2021bvmatch}, \cite{du2020dh3d}. In contrast, our approach streamlines this process by directly estimating the relative pose from the matched filter output.  

LPR methods are mostly validated on widely recognised urban driving datasets like Oxford RobotCar \cite{barnes2020oxford}, KITTI \cite{geiger2013vision}, MulRan \cite{kim2020mulran}, and NCLT \cite{carlevaris2016university}, where they demonstrate high performance. While these datasets are useful for benchmarking, they lack environmental diversity, as they are predominantly captured in structured, urban on-road settings. This focus on urban environments presents another limitation: LPR methods may overfit urban characteristics, such as regular geometric features, sparse occupancy and grid-like street layouts. Consequently, their performance will likely degrade in natural environments with irregular geometric features, seasonally varied vegetation and densely occupied space, as observed in \cite{knights2023geoadapt}. Therefore, evaluating LPR methods in unstructured natural environments like WildPlaces\cite{knights2023wild} is crucial for assessing their generalisation capabilities and promoting the development of robust, environment-invariant place descriptors. 

In this paper, we aim to address these challenges in the field with the following contributions: 
\begin{enumerate}
    \item A two-stage architecture utilising low and high-resolution Birds Eye View (BEV) descriptors for efficient global and local search.
    \item A matching strategy derived from signal processing, using matched filtering to achieve rotation and translation invariant LiDAR place recognition.
    \item Direct relative pose estimation based on matched filtering outputs enabling precise relocalisation 
    \item Extensive experiments on urban and natural environments consistently demonstrate SoTA performance in place recognition and pose estimation, with up to 15\% higher recall than previous SoTA.  
\end{enumerate}

\section{Related Works}
\label{sec:relatedworks}
In this section, we review the current literature to highlight the novelty and contributions of our method, situating it within the broader LPR research landscape. Sections \ref{subsec:handcraftmethods} and \ref{subsec:learnmethods} discuss LPR methods based on handcrafted and learned techniques, respectively. We further explore LPR methods that have been evaluated in unstructured environments in Section \ref{subsec:LPR wild}. Lastly, in Section \ref{subsec:matchedfilter}, we examine the use of correlation-based matching strategies which are similar to our approach using matched filtering.

\subsection{Handcrafted methods for LPR }
\label{subsec:handcraftmethods}
A primary goal of place recognition is to generate descriptors that uniquely and efficiently represent a location. These descriptors can either describe the entire scene globally or extract local distinctive features. Handcrafted methods achieve this by encoding geometric information such as density, height, orientation, depth, or intensity from LiDAR scans. For instance, 3D local descriptors like ISHOT~\cite{guo2019local} store histograms of surface normals, orientation and intensity; while 3D global descriptors like M2DP \cite{he2016m2dp} store signatures of intensity from 2D sliced projections.

Some methods also transformed 3D point clouds into 2D projections with BEV or spherical range images. For example, scan context-based methods \cite{kim2018scan}, \cite{kim2021scan} use polar conversion of BEV as a global place descriptor; while LiDAR Iris \cite{wang2020lidar} creates a binary encoding of BEV polar images with Log-Gabor filters to extract distinctive features. These BEV-based descriptors use circular shifting or rotational search to achieve rotational invariance, similar to our method. Conversely, BVmatch \cite{luo2021bvmatch} applies Log-Gabor filters with a maximum index map to encode orientation information and achieve direct orientation invariance. 

\subsection{Learning-based methods for LPR}
\label{subsec:learnmethods}
The rise of deep learning has also influenced more data-driven approaches to LPR. PointNetVLAD \cite{uy2018pointnetvlad} was a pioneering method, utilising triplet and quadruplet loss functions to train on raw 3D point cloud data end-to-end. This approach was further refined by PCAN \cite{zhang2019pcan}, which introduced an attention layer to better aggregate task-relevant features, and by LPDnet \cite{liu2019lpd}, which employed a graph network for adaptive local feature extraction and neighbourhood aggregation. Similarly, DH3D \cite{du2020dh3d} uses a Siamese network with point convolution to detect and describe local features, which are then aggregated into global descriptors. 

Other deep learning methods have utilised discretised point clouds processed through 3D CNNs. Notable examples include MinkLoc3D \cite{komorowski2022improving}, which incorporates a feature pyramid network, TransLoc3D \cite{xu2021transloc3d}, which employs a transformer network with multiple scale reception and attention mechanisms, and LoGG3D Net \cite{vidanapathirana2022logg3d}, which integrates a sparse U-Net architecture with consistency loss to improve performance.

Despite the high performance of these methods on standard urban datasets like Oxford RobotCar \cite{barnes2020oxford}, KITTI \cite{geiger2013vision}, MulRan \cite{kim2020mulran}, and NCLT \cite{carlevaris2016university}, their generalisation to unstructured environments remains a significant challenge as observed in \cite{knights2023geoadapt}.

\subsection{LPR in Natural Environments }
\label{subsec:LPR wild}
Place recognition within complex unstructured natural environments is crucial for deploying autonomous systems in real-world scenarios and some recent LPR methods have focused on this challenge. For instance, \cite{ou2023place} introduces a handcrafted global descriptor for unstructured orchards, showing generalisation on the Kitti dataset \cite{geiger2013vision}. Since the orchard dataset and reproducible code have not yet been publicly released, we omitted a comparison to this work in our paper. 

On the other hand, the recent release of the WildPlaces dataset \cite{knights2023wild} facilitated LPR evaluation in two natural unstructured environments: Venman and Karawatha. The handcrafted descriptor in BTC \cite{yuan2024btc} is tested on WildPlaces and has high performance in Venman, but the performance significantly degrades in Karawatha. Moreover, the test time adaptation approach in GeoAdapt\cite{knights2023geoadapt} has better performance than standard approaches when trained on urban datasets and tested on WildPlaces, though it still underperforms compared to direct training on WildPlaces. In contrast, our work demonstrates consistent SoTA place recognition performance on the WildPlaces dataset with minimal parameter tuning for urban to natural environments.

\subsection{Correlation based Matching}
\label{subsec:matchedfilter}
Cross-correlation is the underlying technique for matched filters in signal processing. Correlation-based methods have also been applied in autonomous navigation, particularly in scan matching for aligning 2D laser scans, as demonstrated in~\cite{olson2009real} and \cite{konecny2016novel}. In LPR, the Radon Sinogram (RING) method~\cite{lu2022one} uses circular cross-correlation for place recognition and orientation estimation, with RING++ \cite{xu2023ring++} extending this to include 2D cross-correlation for translation estimation. Learning-based LPR methods have also incorporated correlation into their framework. For example, DiSCO \cite{xu2021disco} and OverlapNet \cite{chen2021overlapnet} use correlation for orientation estimation, while DeepRING \cite{lu2023deepring} employs it as a distance vector.

In comparison, our method directly applies cross-correlation with the BEV-transformed query and reference set using matched filtering, enabling both place recognition and relative pose estimation. These handcrafted descriptors are also generated with minimal fine-tuning across diverse environments.

\section {Methodology}

This section outlines our LPR methodology, beginning with an explanation of our two-stage matching architecture in Section \ref{subsec:2stage arch}. Next, we discuss the BEV descriptor generation in Section \ref{subsec:bev desc} and the matched filtering search process in Section \ref{subsec:match filt method}, which are the key components that enable the effectiveness of our method. Finally, we describe the pose estimation calculations using relative shift based on the matched filter output in Section \ref{subsec:pose est method}.

\subsection{Two stage matching architecture}
\label{subsec:2stage arch}
Our method employs a two-stage hierarchical search to estimate query pose: a low resolution global search followed by high resolution local search, as illustrated in Figure~\ref{fig:1.1}. While other methods employ a hierarchical approach, we use the same high and low-resolution descriptors with a matched filter for both place recognition and pose estimation. 

Initially, we construct a lookup grid of low-resolution reference BEV descriptors. In the first global search phase, a low-resolution query BEV descriptor is matched against the reference grid across \(k\) rotation increments (Figure~\ref{fig:1.1} top left). The top \(n\) matches from this phase are obtained based on the maximum correlation.

In the subsequent local search phase, high-resolution query and top \(n\) reference descriptors are matched again, across \(k\) rotations (Figure~\ref{fig:1.1} top right). The best match from this second stage is then used to evaluate the relative shift and estimate the query's pose (Figure~\ref{fig:1.1} bottom). This two-stage architecture, combined with matched filtering enables roto-translation invariant place recognition and direct pose estimation.

\begin{figure}[!t]
    \vspace{0.2cm}
    \centering
    \includegraphics[width=1\columnwidth]{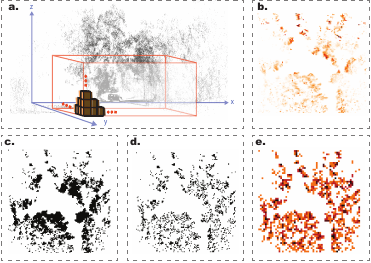}
    \vspace{-0.3cm}
    \caption{Illustrated BEV descriptor generation process with a scan from the WildPlaces Dataset \cite{knights2023wild}: \textbf{a.)} Point cloud data cropped by a 3D window and voxelised. \textbf{b.)} BEV image created from a height density map. \textbf{c.)} Thresholded height density map forming a BEV descriptor indicating occupancy. \textbf{d.)} BEV descriptor after randomly downsampling patches. \textbf{e.)} Lower resolution BEV descriptor after average pooling.}
    \label{fig:3.1}
    \vspace{-0.5cm}
\end{figure}

\subsection{BEV Descriptors}
\label{subsec:bev desc}
A BEV descriptor is generated from the LiDAR scans to uniquely represent each place. A point cloud from a single LiDAR scan is defined by \( P = \{(x_i, y_i, z_i) \mid i = 1, 2, \ldots, N_p \} \), where  \((x_i, y_i, z_i)\) represents the coordinates of a point in 3D space as illustrated in Figure~\ref{fig:3.1} panel $a$. To generate a place descriptor, we extract all points within a specified 3D window: \(-w_x \leq x_i \leq w_x\), \(-w_y \leq y_i \leq w_y\), and \(h_1 \leq z_i \leq h_2\). This cropping step ensures that all scans are constrained to a fixed dimension, which is crucial for consistency during matching. 

Next, a voxel grid is applied to the points to divide the 3D space into regular cubic volumes with a size \(v_x\). The discretised point cloud has reduced noise and complexity, facilitating a consistent scene representation. 

The 3D point cloud is then projected to 2D by creating a BEV image for search efficiency and reduced complexity. This is constructed by generating a height density map (HDM), where we calculate the number of points along the z-axis for each $x$-$y$ bin as visualised in Figure \ref{fig:3.1} panel $b$. 

This HDM BEV image is a standard approach used in LPR methods
\cite{xu2021disco}, \cite{xu2023ring++}, \cite{luo2021bvmatch} and we extend our descriptor generation by including occupancy weight, patch downsampling and average pooling. In our descriptor, a threshold is applied to the density map such that cells with a density greater than \(d\) are considered occupied while all other cells are unoccupied with a predefined weighting \(w\) (Figure \ref{fig:3.1} panel $c$), resulting in: 
\vspace{-0.1cm}
\begin{equation}
    \mathbf{B}(i, j) = \mathbf{1}_{\{\mathbf{HDM}(i, j) > d\}} \cdot w
\end{equation}

The BEV descriptors for the reference scans are also further down-sampled so that each \( m \times m \) patch within the scan have at most \( c \) occupied cells. To achieve this, given the locations of all the occupied cells \( P \) in a patch \(\mathbf{B_{\text{patch}}}\), a subset \( P' \) is randomly selected without replacement to include a maximum of \( c \) cells. The selected cells are considered occupied, while all others are marked as unoccupied with a predefined weighting \(w\). The resulting patches \(\mathbf{B'_{\text{patch}}}\) are combined to form the final downsampled descriptor, with \( p, q \) denoting the index of cells within each patches:
\begin{equation}
    P = \{(p, q) \mid \mathbf{B_{\text{patch}}}(p, q) = 1, 0\le p<m,\quad 0\le q<m\}
\end{equation}
\begin{equation}
     P' \subseteq P \quad \text{such that} \quad |P'| = \min(c, |P|)
\end{equation}
\begin{equation}
    \mathbf{B'_{\text{patch}}}(p, q) =
    \begin{cases}
    w & \text{if } (p, q) \in P' \\
    0 & \text{otherwise}.
    \end{cases}
\end{equation}

 This down-sampling process is particularly useful for dense scans in the reference set, which might otherwise correlate with sparse scans and generate false matches. Figure \ref{fig:3.1} panel $d$, illustrates how patch down-sampling maintains the global geometry while reducing the descriptor density.
 
Finally, average pooling is applied to the descriptors such that only the average of each \(u \times u\) patch is stored, where \(p\) and \(q\) are indices for each patch. This generates the low-resolution BEV descriptors \(\mathbf{B_{\text{lowRes}}}\) (Figure \ref{fig:3.1} panel $e$) used for the global search: 
\begin{equation}
\mathbf{B_{\text{lowRes}}}(i, j) = \frac{1}{u^2} \sum_{p=0}^{u-1} \sum_{q=0}^{u-1} \mathbf{B}(i \cdot u + p, j \cdot u + q).
\end{equation}

\subsection{Matched Filtering}
\label{subsec:match filt method}
Each search, from the two-stage architecture in Figure \ref{fig:1.1}, is conducted with a matched filter. In this process, the centre of each rotated input query \(\mathbf{Q_{rot}}\) is shifted across each point in the reference grid \(\mathbf{R}\) and correlated.  The correlation function calculates the sum of products to measure the query similarity to the reference at every translation offset and rotation increment, defined by:
\begin{equation}
\mathbf{Q}*\mathbf{R}(i_r, j_r, \theta) = \sum_{i_q} \sum_{j_q} \mathbf{Q}(i_q, j_q, \theta) \cdot \mathbf{R}(i_q + i_r, j_q + j_r,0).
\end{equation}

Correlation is equivalent to convolution with a vertically and horizontally reversed query representing the kernel. As a result, the matched filter is represented as a pointwise multiplication in the frequency domain according to the convolution theorem. Hence, the Fast Fourier Transform (FFT) algorithm is used to convert the matched filter inputs into the frequency domain and efficiently compute the filter outputs.

\subsection{Pose Estimation}
\label{subsec:pose est method}
The matched filter output represents a distribution of match similarity across translation and rotation increments \(i_r, j_r, \theta\).  Therefore, the maximum of this output corresponds to the position of the best match within the reference grid and the relative yaw variation between reference and query, as described by:
\begin{equation}
    (i_\text{match},j_\text{match}, \theta_\text{match}) = \argmax_{i_r,j_r, \theta} (\mathbf{Q}*\mathbf{R}(i_r, j_r,\theta)).
\end{equation}

If the maximum correlation position is the centre of a scan in the reference grid, this suggests that the query scan was captured at the same location as the matched reference. However, in realistic datasets, position shifts often occur between the query and reference scans. In such cases, the matched filter output reflects this shift by having the match position offset from the center by the shift amount \(x_{\text{shift}}, y_{\text{shift}}\). By combining this relative shift information with the matched reference pose \(x_r, y_r, \theta_r\), the precise location of the query scan can be estimated with: 
\begin{equation}
\begin{pmatrix}
x_q \\
y_q
\end{pmatrix}
=
\begin{pmatrix}
x_r \\
y_r
\end{pmatrix}
+
\begin{pmatrix}
\cos(\theta_r) & -\sin(\theta_r) \\
\sin(\theta_r) & \cos(\theta_r)
\end{pmatrix}
\begin{pmatrix}
x_{\text{shift}} \\
y_{\text{shift}}
\end{pmatrix}.
\label{Eq:pose}
\end{equation}

\section{Experimental Setup}
\label{sec:experimentalsetup}

\begin{figure}[t]
    \vspace{0.2cm}
    \centering
    \includegraphics[width=0.8\columnwidth]{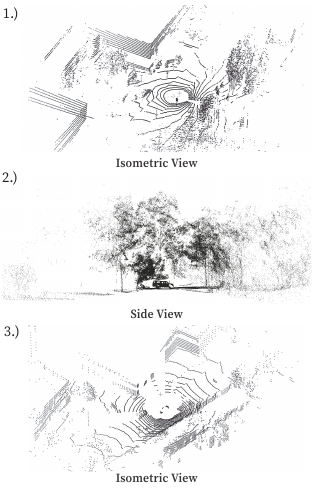}
    \vspace{-0.4cm}
    \caption{Sample LiDAR scans from three different datasets: \textbf{1.)} NCLT Dataset scan using Velodyne 32E-HDL LiDAR sensor captured in an urban driving environment from Michigan \cite{carlevaris2016university}.
    \textbf{2.)} WildPlaces Dataset scan using Velodyne 16-VPL LiDAR sensor captured in unstructured natural environments from Brisbane \cite{knights2023wild}. \textbf{3.)} Oxford Radar Dataset scan using Velodyne32-VPL LiDAR sensor captured in an urban driving environment from Oxford \cite{barnes2020oxford}.}
    \label{fig:4.1}
    \vspace{-0.5cm}
\end{figure}

\subsection{Datasets}
We conducted experiments on three long-term, large-scale datasets from diverse structured and unstructured environments: NCLT \cite{carlevaris2016university},  WildPlaces \cite{knights2023wild} and Oxford Radar RobotCar \cite{barnes2020oxford}, demonstrated in Figure~\ref{fig:4.1}. 

1.) NCLT~\cite{carlevaris2016university} is a long-term urban driving dataset collected on the University of Michigan campus comprising 27 traversals over 15 months. It includes indoor and outdoor environments, multiple seasons, different times of day and various routes captured using a Velodyne HDL 32E LiDAR. The dataset also provides highly accurate ground truth collected using a NovAtel DL-4 with RTK GPS. Using 2 reference traverses from 2013 and 9 traverses from 2012, we test our method on NCLT following the implementation in BVmatch \cite{luo2021bvmatch}, with query scans extracted every 10m. 

2.) WildPlaces~\cite{knights2023wild} is a large-scale dataset with 8 traverses collected in natural, unstructured outdoor environments from Venman and Karawatha in Brisbane, Australia. It covers 33 kilometres in various routes across two different environments (Venman and Karawatha) over 14 months. The scans were collected using a Velodyne VPL 16 LiDAR scanner with a 120-meter range, mounted to a rotating brushless DC motor to achieve 120\textdegree\ vertical FoV. This dataset's submap scans and ground truth positioning information were obtained using WildCat SLAM which integrated the IMU measurements from a Microstrain 3DMCV5-25 9-DoF sensor and the raw LiDAR measurements. We compare our method to the baselines provided with this dataset using the open-source code to determine the query evaluation split for their 8 sequences.

3.) The Oxford Radar RobotCar dataset~\cite{barnes2020oxford} is a radar extension on the popular Oxford Robot Car collected in 2019. It consists of 32 traversals through urban central Oxford capturing various weather, traffic and lighting conditions. The LiDAR scans were obtained using two Velodyne HDL-32E sensors and we arbitrarily picked the left sensor for this investigation. It also comes with ground truth positioning information from NovAtel SPAN-CPT ALIGN INS/GPS sensor. Following BVmatch \cite{luo2021bvmatch}, we evaluate Oxford Radar query scans from every 10m using our method with 2 reference traverses and 5 query traverses.

\subsection{Evaluation Metrics}
In this work, we employ a combination of place recognition and pose estimation metrics, following \cite{xu2023ring++}, \cite{luo2021bvmatch}, \cite{knights2023wild}, to evaluate our method. We calculate recall @1 for all datasets by determining the percentage of true positives or estimated locations within a specified distance threshold. For WildPlaces, we use a threshold of 3$m$, and for NCLT and Oxford Radar, we use a threshold of 25$m$, following the comparison methods. This is calculated using:
\begin{equation}
    \text{Recall @1} = \frac{TP}{TP+FP}.
\end{equation}

Since our method predicts a 3DOF pose estimate, we evaluate the rotation and translation errors of the positive place recognition pairs based on recall @1. The relative translational error (RTE) is calculated as the Euclidean distance between the ground truth 2D position \(\rho_{\text{GT}}\) and the estimated position \(\rho_\text{est}\): \(\text{RTE} = ||\rho_{\text{GT}} - \rho_\text{est}||\).

The relative rotation error (RRE) is the absolute difference between the ground truth yaw \(\gamma_{\text{GT}}\) and the estimated yaw \(\gamma_\text{est}\): \(\text{RRE} = |\gamma_{\text{GT}} - \gamma_\text{est}|\).

An additional success rate criteria is measured for pose estimation, following BVmatch \cite{luo2021bvmatch}. The success rate is defined as the percentage of pose estimates that have less than 2$m$ RTE and 5$^{\circ}$ RRE. This is formulated as:
\begin{equation}
    \text{SR} = \frac{TP_{RTE<2m\ \land \text{ } RRE<5^{\circ}}}{TP}.
\end{equation}

\subsection{Implementation Details}
We selected and fine-tuned hyperparameters for our method with ablation in Section \ref{subsec:ablate} to ensure consistency across experiments and datasets.Key parameters including 120$\times$120 high-resolution descriptor size, top 2 matches (\(n\)), 2$\times$2 average pooling window (\(u\)), $-0.15$ unoccupied cell weighting (\(w\)), 10$^\circ$ rotation increment (\(k\)), 10x10 patch window for downsampling (\(m\)) and 20 maximum occupied cells in a patch (\(c\)), were standardised across all experiments. Additionally, we collected scans at 2-meter intervals to maintain an even distribution for the reference set.

To adapt to varying environmental conditions, we also fine-tuned specific parameters. To capture sufficient detail from dense point clouds, we measured the average number of points in BEV descriptors with 1\(m^3\) voxel size for each dataset. Voxel sizes (\(v_x\)) were then adjusted following an inverse linear relationship with 1360, 876 and 275 average point density, resulting in 0.3\(m^3\), 0.75\(m^3\) and 1.3\(m^3\) for WildPlaces, NCLT and Oxford Radar, respectively. Additionally, a threshold density ($d$) of 1 was used for urban and 2 for natural environments, ensuring consistent descriptor density. Furthermore, each dataset's height and intensity thresholds were adjusted to remove the ground plane noise. In the WildPlaces dataset, the height was cropped between $1m$-$3m$ to remove less descriptive information from the tree foliage and extract the shape of the cleared path.

\section{Results and Discussion}
\label{sec:results}
\vspace{-0.1cm}
\begin{table}[!h]
\caption{Average Recall @1 on Wildplaces Datatset}
\vspace{-0.2cm}
\centering
\begin{tabular}{c|ccc}
\toprule
Method & Venman & Karawatha & Average\\
\midrule
Scan Context \cite{kim2018scan} & 33.98 & 38.44 & 36.21 \\
TransLoc3D \cite{xu2021transloc3d} & 50.24 & 46.08 & 48.16 \\
MinkLoc3Dv2 \cite{komorowski2022improving} & 75.77 & 67.82 & 71.80 \\
LoGG3D-Net \cite{vidanapathirana2022logg3d} & 79.84 & 74.67 & 77.26 \\
Ours & \textbf{94.46} & \textbf{90.50} & \textbf{92.48} \\
\bottomrule
\end{tabular}
\label{tab:Wildplaces_performance}
\vspace{-0.3cm}
\end{table}

\begin{table}[!h]
\caption{Average Recall @1 on NCLT \& Oxford Radar}
\vspace{-0.2cm}
\centering
\begin{tabular}{c|cc}
\toprule
Method & Oxford Radar & NCLT \\
\midrule
M2DP \cite{he2016m2dp} & 50.9 & 36.2  \\
PN-VLAD \cite{uy2018pointnetvlad} & 86.6 & 62.8  \\
PCAN \cite{zhang2019pcan} & 80.3 & 59.0  \\
LPD-Net \cite{liu2019lpd} & 90.0 & 72.5  \\
DH3D \cite{du2020dh3d} & 78.2 & 59.4  \\
BVMatch \cite{luo2021bvmatch} & 93.9 & 83.6  \\
Ours & \textbf{94.6} & \textbf{92.9} \\
\bottomrule
\end{tabular}
\label{tab:oxford nclt table}
\vspace{-0.2cm}
\end{table}

This section presents the performance of our method across three datasets: Section \ref{subsec:placeRecNatural} evaluates performance in natural environments following the implementation in WildPlaces \cite{knights2023wild}, while Section \ref{subsec:placeRecUrban} focuses on urban environments (NCLT and Oxford Radar RobotCar) following the implementation in BVmatch \cite{luo2021bvmatch}. Section \ref{subsec:placeRecCorr} compares our method with other correlation-based approaches. We assess our pose estimation capabilities in Section \ref{subsec:poseEstNCLT} and present an ablation study on hyperparameters in Section \ref{subsec:ablate}. Finally, Section \ref{subsec: runtime} provides a runtime evaluation of our descriptor generation and search architecture.

\subsection{Place Recognition in Natural Environment}
\label{subsec:placeRecNatural}

 In the WildPlaces dataset, we conducted 24 experiments, comparing every combination of reference and query from 4 traverses in both Karawatha and Venman. The average recall of our method compared to \cite{kim2018scan}, \cite{xu2021transloc3d}, \cite{komorowski2022improving}, \cite{vidanapathirana2022logg3d} is presented in Table \ref{tab:Wildplaces_performance}. We observe large performance improvements, with $\sim$15\% higher recall @1 for both Venman and Karawatha sequences on average. 
 
\begin{figure}[!t]
    \vspace{0.2cm}
    \centering
    \includegraphics[width=0.95\columnwidth]{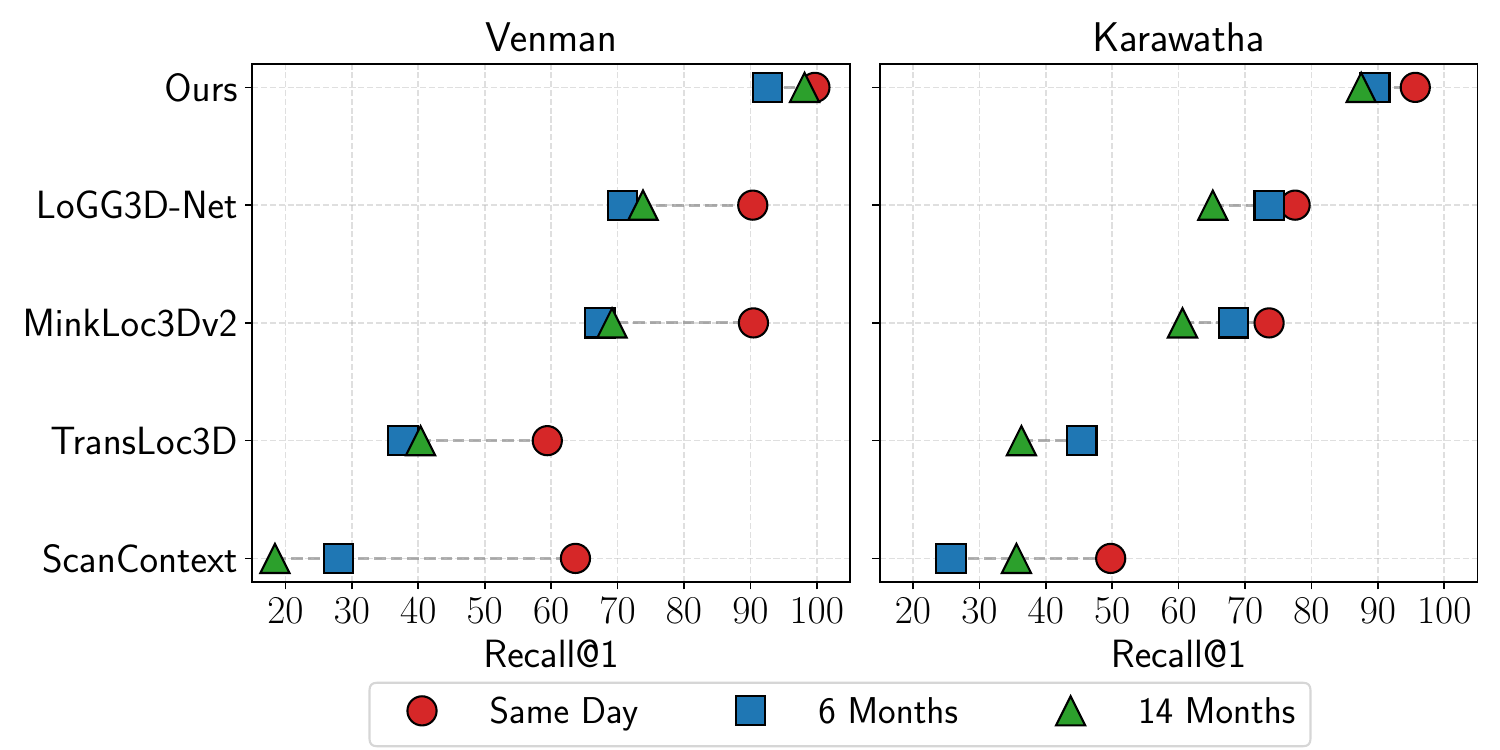}
    \vspace{-0.3cm}
    \caption{Long Term Place Recognition Performance. Examination of LPR performance for query collected from day one, tested on references from the same day, 6 months later and 14 months later using 5 different methods, including ours.}
    \label{fig:wildplacesOverTime}
    \vspace{-0.6cm}
\end{figure}

 Since the WildPlaces dataset was collected over 14 months, we also evaluated long-term place recognition to analyse the challenges arising from long-term changes in a dense forest. Using sequence 2 collected on June 2021 as a query, we measured recall @1 on sequence 1 (same day, June 2021), sequence 3 (6 months later, December 2021) and sequence 4 (14 months later, August 2022).  These long-term sequences faced different challenges: winter-to-summer seasonal change over 6 months and longer-term variations over a 14 months. Following the trend of the comparison methods, our method also exhibited the highest performance on the same day as illustrated in Figure \ref{fig:wildplacesOverTime}. We observed a 7\% drop in recall after 6 months and an 8\% drop after 14 months, compared to 20\% and 12\% drop in the best baseline for Venman and Karawatha, respectively. Our long-term place recognition at 14 months also outperforms the same-day performance of all comparison methods. 
 
 A possible reason for the improved performance is the additional BEV descriptor processing and voxel size fine-tuning which allowed our method to handle denser point clouds. 

\subsection{Place Recognition in Urban Environments}
\label{subsec:placeRecUrban}
For the urban NCLT and Oxford radar datasets, we compare to a combination of handcrafted features and deep learning techniques ~\cite{he2016m2dp, uy2018pointnetvlad, zhang2019pcan, liu2019lpd, du2020dh3d, luo2021bvmatch}, following the results presented in BVmatch~\cite{luo2021bvmatch}. Using the performance metric recall @1, we outperform the baselines on both datasets with significant improvement, $\sim$10\%, in NCLT and 0.7\% in Oxford Radar, as presented in Table \ref{tab:oxford nclt table}. We note that although the baselines exhibit performance variations of around 10\% across the two urban datasets, which were captured with the same sensor in similar environments (as illustrated in Figure \ref{fig:4.1}), our method demonstrates consistently high performance across both datasets. 

\subsection{Place Recognition with Correlation-based Methods}
\label{subsec:placeRecCorr}
We also compare the place recognition performance of our method with other correlation-based matching methods, specifically RING++ \cite{xu2023ring++}, which uses handcrafted features, and DeepRING \cite{lu2023deepring}, which employs learned feature extraction. While we initially aimed to evaluate these methods using the BVMatch evaluation split, we were unable to reproduce the reported results. Consequently, we reran our experiments using the evaluation splits from their respective papers and compared our performance to their reported results.

RING++ reports a recall @1 of 73.2\% compared to 90.8\% for our method. This was tested on the NCLT dataset using a revisit threshold of 10 meters, with the reference sequence ``2012-02-04'' sampled every 20 meters and the query sequence ``2012-08-20'' sampled every 5 meters. DeepRING, utilising the same reference sequence and sampling intervals but with the query sequence ``2012-03-17,'' achieves a recall @1 of 85.9\%  whereas our method achieves 88.0\%. Without any changes to our hyperparameters, our method outperforms both RING++ and DeepRING in terms of place recognition recall.

\subsection{Pose Estimation}
\label{subsec:poseEstNCLT}

\begin{table}[t]
\vspace{0.2cm}
\caption{Pose Estimation Results on NCLT}
\centering
\begin{tabular}{l|cc|cc|c}
\toprule
\multirow{2}{*}{Method} & \multicolumn{2}{c|}{RTE (m)} & \multicolumn{2}{c|}{RRE (deg)} & \multirow{2}{*}{SR (\%)} \\
                        & Mean & Std. & Mean & Std. &  \\
\midrule
SIFT \cite{lowe2004distinctive} & 0.69 & 0.46 & 1.49 & 1.23 & 53.3 \\
DH3D \cite{du2020dh3d} & 1.16 & 0.54 & 3.43 & 1.06 & 14.8 \\
OverlapNet \cite{chen2021overlapnet} & - & - & 2.43 & 1.42 & 15.4 \\
BVMatch \cite{luo2021bvmatch} & 0.57 & 0.38 & \textbf{1.08} & \textbf{1.00} & 94.5 \\
Ours & \textbf{0.48} & \textbf{0.29} & 1.43 & 1.07 & \textbf{97.7} \\
\bottomrule
\end{tabular}
\label{tab:pose est}
\vspace{-0.3cm}
\end{table}

To evaluate our pose estimation method, we compared it against the results reported in BVmatch \cite{luo2021bvmatch}. Performance was assessed using RTE, RRE, and SR metrics on a single reference-query pair from the NCLT dataset, as shown in Table \ref{tab:pose est}. Our method achieved 3.5\% more successful matches and demonstrated 0.09m lower mean and deviation in RTE for the correct poses. However, due to the coarse rotation search used in our approach, BVmatch, which incorporates ICP refinement, achieved a 0.35$^\circ$ lower mean and a 0.07$^\circ$ lower deviation for RRE. Ablation of the hyperparameters in Figure \ref{fig:ablation} shows our performance can be improved with finer rotation increments, albeit at the cost of increased run time.

\begin{figure}[t]
    \vspace{0.2cm}
    \centering
    \includegraphics[width=0.7\columnwidth]{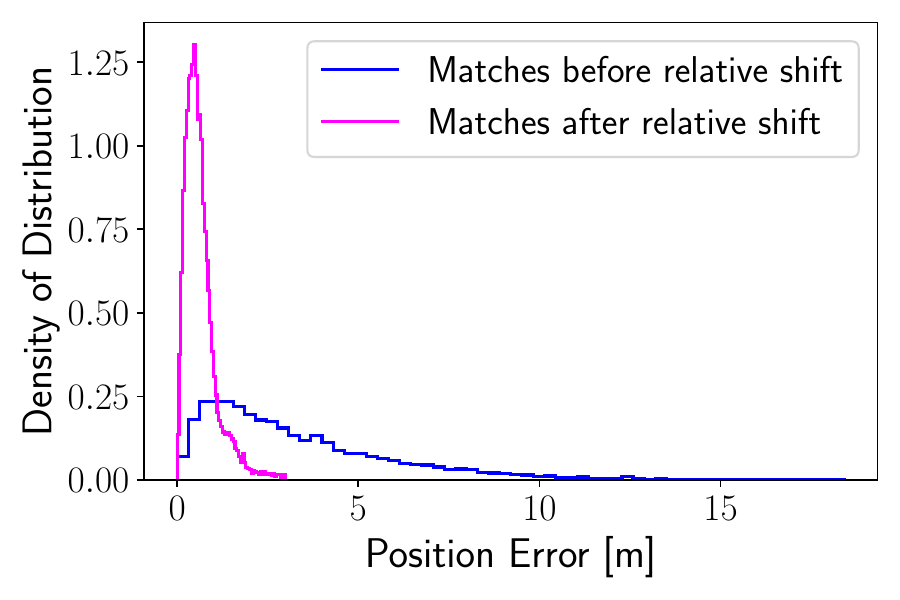}
    \vspace{-0.4cm}
    \caption{Impact of the relative shift on position estimation for the complete WildPlaces dataset: The position error density distribution of correct LPR matches with and without relative pose correction demonstrates a lower mean error and smaller deviation after correction.}
    \label{fig:errorRelativeShift}
\end{figure}

Unlike the comparison methods, the primary advantage of our pose estimation approach lies in its ability to directly estimate relative shifts from the matched filter output, which is then used to correct the reference pose. This advantage is evident in Figure \ref{fig:errorRelativeShift}, which shows the distribution of position errors for successful matches in the WildPlaces dataset. After applying relative shift correction, the mean and standard deviation of pose errors were significantly reduced. Consequently, our roto-translation invariant place search not only enhances place recognition but also facilitates direct and accurate pose estimation.

\begin{figure}[!t]
    \centering
    \includegraphics[width=0.95\columnwidth]{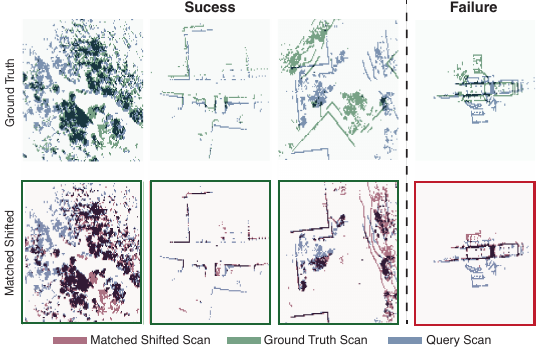}
    \caption{A selection of matched scans with their corresponding ground truth overlaid on the query. The scans were selected from WildPlaces (column 1), Oxford Radar (column 2), and NCLT (columns 3 and 4). A green outline around the first three scans indicates a successful match, while a red outline indicates an incorrect match.}
    \label{fig:sample scans}
    \vspace{-0.4cm}
\end{figure}

The effectiveness of our pose estimation method can also be visualised using BEV descriptors. Figure \ref{fig:sample scans} illustrates a selection of query scans overlaid with ground truth and matched shifted scans, presented in two rows. Our method successfully identifies matches in dense outdoor environments and scenarios with significant translation and rotation variations in the ground truth scans. A failure case is also highlighted from a narrow path within the NCLT dataset, where multiple aliased scans in the reference led to an incorrect match.

\subsection{Parameter Ablation}
\label{subsec:ablate}
\begin{figure}[!t]
\vspace{0.2cm}
    \centering
    \includegraphics[width=0.99\columnwidth]{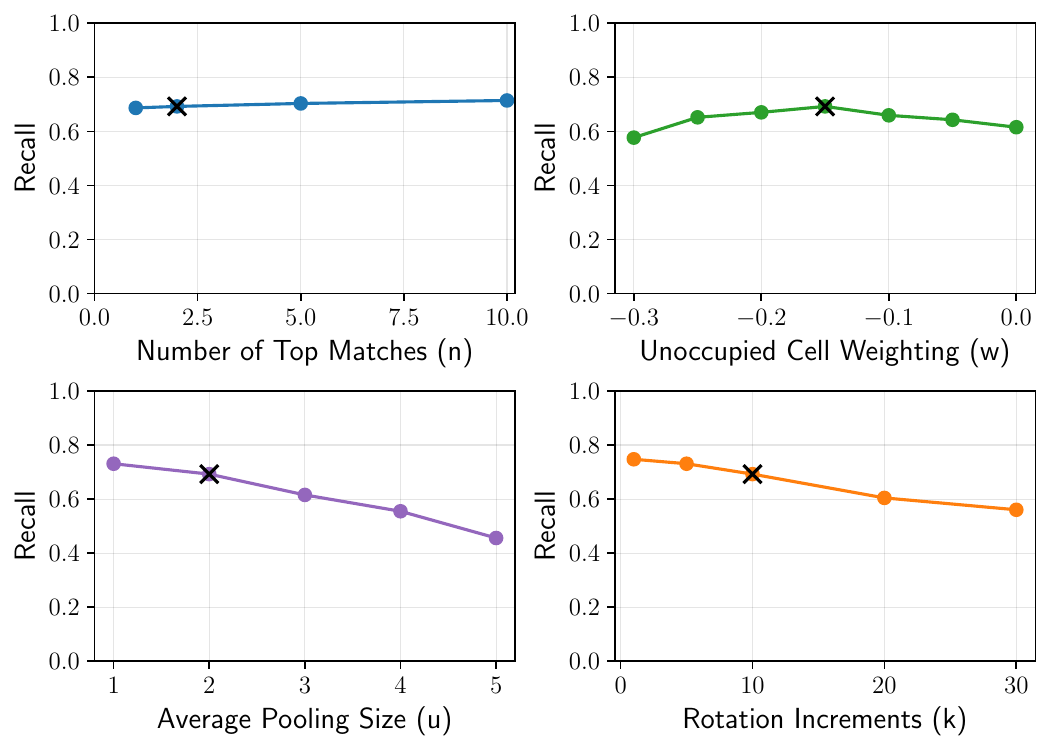}
    \vspace{-0.5cm}
    \caption{Ablation of Parameters. Examination of recall on sequence 2 (reference) and 3 (query) from WildPlaces Venman Dataset while varying: \textbf{Top left -} number of top matches ($n$), \textbf{Top right -} weighting ($w$) for unoccupied cells in the BEV descriptor, \textbf{Bottom left -} average pooling size ($u$), \textbf{Bottom right -} rotation increment ($k$) for query scans. The black x indicates the chosen value for each parameter in this investigation.}
    \label{fig:ablation}
    \vspace{-0.4cm}
\end{figure}

We also conducted an ablation study on the impact of key parameters on place recognition performance. The ablation was implemented using sequence 2 as reference and sequence 3 as query from the WildPlaces Venman dataset, as illustrated in Figure~\ref{fig:ablation}. This shows the performance of our system can be further improved by choosing a larger \(n\) for top matches, a smaller \(k\) for rotation increments, or by removing average pooling. However, the values chosen in our implementation minimise computational costs while maintaining acceptable performance.

\subsection{Runtime Evaluation}
\label{subsec: runtime}

We implemented and evaluated our method on a desktop computer with an NVIDIA GeForce RTX 3090 GPU and an 11th Gen Intel Core i7 processor. The average descriptor generation times were approximately 30ms for Oxford Radar, 40ms for NCLT, and 300ms for WildPlaces. The longer descriptor generation time for WildPlaces is attributed to the denser point clouds with a larger number of data points for processing. Since the descriptor sizes of 60$\times$60 and 120$\times$120 were consistent across all evaluations, the two-stage search time depends on the reference set size. With 1500 reference scans, after subsampling every 2 meters, the average global search time is 460ms, and the local search time is 120ms. Consequently, the overall method, including place recognition and pose estimation, operates at approximately 1Hz with 1500 reference scans.

\section{Conclusions and Future Work}
\label{sec:conclusion}

In this paper, we presented a two-stage, roto-translation invariant LiDAR place recognition method capable of direct relative pose estimation. Our method demonstrated SoTA performance in both urban and unstructured environments, with minimal fine-tuning. One limitation of our method is the assumption of minimal pitch and roll variations, which can impact the accuracy of the BEV projection. Additionally, the runtime could be optimised for longer traverses involving several thousand reference scans. To address these challenges, future work will focus on generating a comprehensive BEV map of the entire reference traverse, reducing search time and enhancing efficiency. This LPR method also holds potential for integration within a SLAM pipeline to enable loop closure detection in future works.

\bibliography{Bibliography}
\end{document}